\documentclass[manuscript]{acmart}
\AtBeginDocument{%
  }

\acmJournal{JACM}
\acmVolume{37}
\acmNumber{4}
\acmArticle{111}
\acmMonth{6}

\usepackage{multirow}

\begin{document}

\title{Position Prediction Self-Supervised Learning for Multimodal Satellite Imagery Semantic Segmentation}

\author{John Waithaka}
\email{jwaithak@andrew.cmu.edu}
\author{Moise Busogi}
\email{mbusogi@andrew.cmu.edu}
\affiliation{%
  \institution{Carnegie Mellon University Africa}
  \city{Kigali}
  \country{Rwanda}
}

\renewcommand{\shortauthors}{Waithaka et al.}

\begin{abstract}
  Semantic segmentation of satellite imagery is crucial for Earth
  observation applications, but remains constrained by limited
  labelled training data. While self-supervised pretraining
  methods like Masked Autoencoders (MAE) have shown promise, they
  focus on reconstruction rather than localisation—a fundamental
  aspect of segmentation tasks. We propose adapting LOCA
  (Location-aware), a position prediction self-supervised learning
  method, for multimodal satellite imagery semantic segmentation.
  Our approach addresses the unique challenges of satellite data
  by extending SatMAE's channel grouping from multispectral to
  multimodal data, enabling effective handling of multiple
  modalities, and introducing same-group attention masking to
  encourage cross-modal interaction during pretraining. The
  method uses relative patch position prediction, encouraging
  spatial reasoning for localisation rather than reconstruction.
  We evaluate our approach on the Sen1Floods11 flood mapping
  dataset, where it significantly outperforms existing
  reconstruction-based self-supervised learning methods for satellite
  imagery. Our results demonstrate that position prediction tasks,
  when properly adapted for multimodal satellite imagery, learn
  representations more effective for satellite image semantic segmentation than
  reconstruction-based approaches. Source code is available at
  \url{https://github.com/johnGachihi/scenic}.
\end{abstract}




\received{6 June 2025}

\maketitle

\section{Introduction}
Satellite imagery is a fundamental data source for Earth observation research, with semantic segmentation being particularly important for analysing this imagery. Semantic segmentation enables, for example, the extraction of flood extent maps, crop cover maps, and forest cover maps for disaster management, food security analysis, and climate research.

While deep learning models have proven effective for semantic segmentation of satellite imagery (for example in \cite{Zhang2022, Li2020}), semantic segmentation remains constrained by limited labelled training data. The pixel-level annotation for semantic segmentation is extremely expensive and time-consuming to obtain \cite{zhou@ade20k, Caron2024loca}, and satellite imagery adds to this challenge due to lower spatial resolution, unfamiliar semantic classes, and the need for domain expertise.

% Pretraining is commonly used to improve model performance when labeled training data is limited. Supervised pretraining, the dominant pretraining technique, typically uses image-level tasks such as classification [C] and image-text alignment [C] on large-scale natural image datasets such as LAION-5B or ImageNet-21K, because, Caron et al., [C] argue, it is cheaper to obtain image-level annotations than pixel-level annotations. However, this approach is suboptimal for satellite imagery semantic segmentation. First, models pretrained on natural images transfer less effectively to satellite imagery tasks than models pretrained on satellite imagery datasets [C, C]. Second, image-level pretraining tasks may be poorly aligned with pixel-level downstream tasks like semantic segmentation [C].

Pretraining is commonly used to improve model performance when
labelled training data is limited. Self-supervised pretraining,
which does not require labelled data, particularly fits
the satellite imagery domain, where, although there is a scarcity
of labelled datasets, there are massive unlabelled satellite
imagery datasets.

Contrastive learning is a prominent self-supervised pretraining
method. It involves matching two different views of the same thing,
generated through separate data augmentation draws or temporal
displacement \cite{Ayush2022Geography-AwareLearning}. However,
Caron et al. \cite{Caron2024loca} find that models pretrained with
contrastive learning do not transfer well to semantic segmentation
tasks. They hypothesise this occurs because contrastive learning
encourages global image-level representation with no need for
spatial reasoning, whereas semantic segmentation is a pixel-level
task that, intuitively, benefits from spatial reasoning.

Masked image modelling, particularly the Masked Autoencoder (MAE)
pretraining scheme \cite{He2022mae}, have been widely explored in
the satellite imagery domain \cite{Cong2022satmae, Reed2023scalemae, Tang2023CrossScaleMae,
Nedungadi2025mmearth, Noman2024Satmaepp}. MAE defines a masked
patch reconstruction task for self-supervised pretraining. This
task encourages spatial reasoning as visible patches in
different spatial positions predict masked patches in other
positions. MAE-based methods, in some prior works, have
outperformed contrastive methods on satellite imagery semantic
segmentation \cite{Reed2023scalemae, Nedungadi2025mmearth}.

Location prediction is a less prominent self-supervised pretraining
method. LOCA (Location-aware) \cite{Caron2024loca}, in particular,
defines a relative location prediction task for self-supervised
learning. More precisely, a query and reference view are sampled
from an input image, and each patch in the query view predicts
its position in the reference view. This task encourages spatial
reasoning for localisation, unlike MAE which encourages spatial
reasoning for reconstruction. Since, segmentation is, in part,
fundamentally a localisation task, we hypothesise that relative
location prediction learns patch representations that are more
effective for semantic segmentation. Further, Caron et al.
\cite{Caron2024loca}, show that LOCA outperforms other self-supervised
methods on various semantic segmentation datasets in the natural
image domain. However, relative location prediction remains
unexplored in the satellite imagery domain.

Satellite imagery has significant differences from natural
imagery. Whereas natural images typically consist only of RGB
bands, satellite images can consist of more bands from a wider
range of the electromagnetic spectrum. Further, since satellite
images are captured by different kinds of Earth observation sensors,
there exist `multimodal' images giving complementary views of
the same geolocations. We adopt LOCA to effectively handle the
multispectral nature of satellite imagery as well as to exploit its
multimodality to improve transfer performance on satellite
imagery semantic segmentation.

In this work, we adapt LOCA for multimodal satellite imagery
semantic segmentation by extending channel grouping to handle
multiple modalities (multispectral imagery, SAR, and DEM) and
introducing same-group attention masking to encourage cross-modal
interaction during pretraining. Evaluation on the Sen1Floods11
\cite{Bonafilia2020Sen1Floods11} flood mapping dataset demonstrates
that our position prediction approach outperforms existing
reconstruction-based self-supervised learning methods for satellite
imagery.

\section{Related Work}
\subsection{Position Prediction for SSL and LOCA}
A relatively unpopular branch of self-supervised learning (SSL) is patch
position prediction. Patch position prediction methods exploit the
spatial context in images to define a pretext task. These tasks involve
predicting the spatial position of patches in an image. Doersch et al.
\cite{doersch@contextpred} sample two patches from the same image and
predict the position of one patch relative to the other. Noroozi and
Favaro \cite{noroozi@jigsaw} divide an image into nonoverlapping 
patches and predict their true positions after they have
been shuffled. Zhai et al. \cite{zhai@pospred}, using vision
transformers, predict the positions of patches given the patches without
positional information (position encoding \cite{vaswani@transformers}).

Our work is based on LOCA \cite{Caron2024loca}. LOCA defines a relative
patch position prediction task. More precisely, a query view and
reference view are sampled from an image, and each patch in the query
view predicts its position relative to the reference view. To this end,
the query view patches attend to the reference view through a single
cross-attention block. To control the difficulty of the task, a
fraction of the reference view is made visible to the query view. Caron
et al. \cite{Caron2024loca} show that LOCA outperforms other SSL
pretraining methods on a number of natural image semantic segmentation
datasets, however, position prediction methods remain underexplored in
the satellite imagery domain.

\subsection{Patch Clustering for Dense SSL}
Ziegler and Asano \cite{ziegler@leopart} use clustering to generate
pseudo-labels for supervising a patch-level classification task. The
cluster assignment is done online using a teacher network (and cluster
prediction by a student network). LOCA \cite{Caron2024loca} uses the
same technique in addition to relative position prediction.

\subsection{Multimodal Pretraining for Satellite Imagery}
Multimodal learning attempts to build AI models that can extract and
relate information from multiple modalities \cite{Xu@MultimodalSurvey, Baltrusaitis@multimodalSurvey}. This is inspired by human perception,
which collects data of different modalities (e.g., visual, auditory)
and uses them complementarily to get a more complete understanding of
an environment. A modality is associated with a certain sensor that
captures a distinct type of data \cite{Xu@MultimodalSurvey}.

In the Earth observation domain, multiple sensors capture different
views of the Earth, each view containing distinct and useful
information. These views are different enough that prior works view
them as different modalities \cite{Guillaume2025Omnisat,
Nedungadi2025mmearth}. There is significant research
interest in how to use multimodal satellite imagery to create more
effective Earth observation solutions \cite{Ghamisi2019}.

In this work, we consider three modalities: multispectral satellite
imagery (MSI), synthetic aperture radar (SAR), and digital elevation
model (DEM). MSI captures reflected or emitted radiation energy from
a range of wavelengths on the electromagnetic spectrum, from visible
light to thermal infrared radiation \cite{WilliamEmery2017}. SAR
images are captured by an active sensor that emits microwave energy
to the earth and measures how much of it is scattered back to the
sensor. SAR images have the benefit of not being affected by the
weather or cloud cover. DEM contains pixel-level surface elevation
data.

Multimodal self-supervised pretraining in the satellite imagery
domain has been studied previously \cite{Nedungadi2025mmearth,
Guillaume2025Omnisat, Han2024msGFM, tseng2025galileolearningglobal}.
Nedungadi et al.
\cite{Nedungadi2025mmearth} build on masked autoencoders (MAE)
\cite{He2022mae} for multimodal image reconstruction given single-modal
input. They achieve this through multiple modality-specific
MAE reconstruction decoders. Han et al. \cite{Han2024msGFM} and Astruc et
al. \cite{Guillaume2025Omnisat} also build on MAE but use
multimodal input for multimodal reconstruction. They achieve this
through multiple modality-specific embedders, a cross-modal encoder, and
multiple modality-specific reconstruction decoders. Recently, Tseng et
al. \cite{tseng2025galileolearningglobal} use a novel `global and local'
cross-modal latent representation reconstruction task for SSL. All prior
work on multimodal self-supervised pretraining in satellite imagery
found use a form of masked image reconstruction. Multimodal
self-supervised pretraining on satellite imagery using position
prediction tasks remains unexplored.

\subsection{Masked Autoencoders for Satellite Imagery}
Masked autoencoders \cite{He2022mae} are ViT-based self-supervised
learners. Following masked language modelling in NLP (e.g. BERT
\cite{Devlin2019}), MAE learns image representations by reconstructing
masked patches of an image given visible patches. MAE has been widely
explored in the satellite imagery domain \cite{Cong2022satmae,
Reed2023scalemae, Noman2024Satmaepp, Nedungadi2025mmearth,
Guillaume2025Omnisat, Han2024msGFM, tseng2024presto}.

MAE, like position prediction, encourages spatial reasoning and thus
learns image representations suitable for semantic segmentation
transfer. However, MAE uses spatial reasoning for reconstruction,
whereas position prediction tasks use it for localisation. Since,
semantic segmentation is, in part, fundamentally a localisation
task, we argue that position prediction tasks will learn representations
more suitable for semantic segmentation transfer. We compare transfer
performance of MAE with our work.

\section{Methodology}
Our work builds on LOCA \cite{Caron2024loca}, adopting it for
multimodal satellite imagery. We detail our adaptations as well as
what is borrowed from LOCA.

\paragraph{Sampling query and reference views}
Multimodal image pairs are concatenated along the channel dimension
to form a single input image $x$. Following LOCA, we sample a query
view $x_q$ and reference view $x_{ref}$ from $x$, then apply
independent random augmentations (i.e., flipping, cropping, rescaling)
to each view. To maximise overlap between corresponding query and
reference views while ensuring queries represent local image regions,
reference views are sampled to cover a large area of the original
image and query views to cover small portions of the original image.
Following LOCA, we sample 10 query views per reference view.

\paragraph{Query and reference patch position correspondence}
Query and reference views are divided into nonoverlapping $P \times
P$ patches. Each query view thus yields patches $x_q^i$ for $i \in
\{1, ..., N_q\}$, where $N_q = \lfloor H_q / P\rfloor \times \lfloor W_q / P\rfloor$
and $H_q \times W_q$ is the query resolution. We use $H_q = W_q = 96$
and $P = 16$ yielding $N_q = 36$ patches per query. Similarly, the
reference view yields patches $x_{ref}^j$ for $j \in \{1,..., N_{ref}\}$.
We use $H_{ref} = W_{ref} = 224$ and $N_{ref} = 196$. To maintain
spatial position correspondence across augmentations, we track each
patch's original position. This allows us to define a mapping
function $h(i) = j$ that identifies the reference patch $x_{ref}^j$
with the greatest overlap to query patch $x_q^i$.

\paragraph{Channel grouping}
Both query and reference patches have $C$ channels:
$x_q^i, x_{ref}^j \in \mathbb{R}^{P \times P \times C}$.
Following SatMAE \cite{Cong2022satmae}, we partition these channels
into $G$ channel groups of $g$ channels each. Each group is processed
by a separate patch embedding to produce token sequences
$S_q^g \in \mathbb{R}^{N_q \times d}$ and
$S_{ref}^g \in \mathbb{R}^{N_{ref} \times d}$ for $g \in \{1,...,G\}$.
These sequences are concatenated along the sequence dimension,
yielding $S_q \in \mathbb{R}^{GN_q \times d}$ and
$S_{ref} \in \mathbb{R}^{GN_{ref} \times d}$. Channel grouping gives
us the flexibility to form token sequences, say, from a mixture of
modalities or separately for each modality. We perform ablations on
different channel group settings.

\paragraph{Group encoding}
Following SatMAE \cite{Cong2022satmae}, we apply group and positional
encodings to retain spatial and channel group information. Each token
receives a group encoding $GE_g \in \mathbb{R}^d_{GE}$ and positional
encoding $PE_i \in \mathbb{R}^{d_{PE}}$ where $d_{GE} + d_{PE} = d$.
These encodings are concatenated and added to the corresponding
tokens in both $S_q$ and $S_{ref}$.

\paragraph{Group sampling}
Channel grouping increases the sequence length from $N_q$ to $GN_q$
tokens for queries (and $N_{ref}$ to $GN_{ref}$ for references). To
maintain computational efficiency, we sample one token per spatial
position (each position has $G$ tokens for each group), preserving
the original sequence length $N_q$ and $N_{ref}$. We sample
uniformly across channel groups to ensure balanced
representation, yielding $S^\prime_q \in \mathbb{R}^{N_q \times d}$
and $S^\prime_{ref} \in \mathbb{R}^{N_{ref} \times d}$.

\paragraph{Transformer self-attention encoder blocks}
The sampled sequences $S^\prime_q$ and $S^\prime_{ref}$ are processed
independently through transformer encoder blocks, yielding query and
reference representations $Z_q \in \mathbb{R}^{N_q \times d}$
and $Z_{ref} \in \mathbb{R}^{N_{ref} \times d}$.

\paragraph{Query-reference interaction}
Caron et al. \cite{Caron2024loca} claim that to solve the relative
patch position prediction task, query patch representations must
attend to the corresponding reference patch representations. Following
LOCA \cite{Caron2024loca}, we implement this using a single
cross-attention block whose queries are computed from $Z_q$ and
keys/values from $Z_{ref}$, yielding output $U \in \mathbb{R}^{N_q \times d}$.

\paragraph{Patch position prediction}
To learn spatial relationships without annotations, we follow LOCA
\cite{Caron2024loca} and solve a relative patch position prediction
task. This is formulated as a $N_{ref}$-way classification task where
each query patch predicts its corresponding reference patch position
from among the $N_{ref}$ positions.
In particular, a classification layer processes the query patch
representations $U$ to output the position predictions
$O \in \mathbb{R}^{N_{ref} \times N_q}$ for each query patch. We
minimise the loss
\begin{equation}
    \frac{1}{|\Omega|} \sum_{j \in \Omega} \ell(O_j, h(j)) \label{pos-pred-loss}
\end{equation}

where $\Omega$ is the set of query patches with a corresponding
patch position in the reference view and $\ell$ is the softmax
cross-entropy loss.

\paragraph{Same-group attention masking} To encourage cross-group
and cross-modal interaction we prevent patches within the same group
from attending to each other in both self-attention and
cross-attention blocks. This encourages the model to form
representations based on information from different groups and
modalities rather than over-relying on patches from the same group.
In particular, we define a binary mask $M$ where $M_{i,j} = 0$ if
patches $i$ and $j$ belong to the same group, and $M_{i,j} = 1$
otherwise. This masking is applied to:
\begin{itemize}
    \item Self-attention: preventing within-group attention among
    query patches or among reference patches.
    \item Cross-attention: preventing query patches from attending to
    reference patches in the same group
\end{itemize}
The masked attention is computed as:
\begin{displaymath}
    E = \text{softmax}(\frac{QK^\top}{\sqrt{d}} \odot M)V
\end{displaymath}
where $K$, $Q$ and $V$ are the standard key, query and value attention
maatrices, and $\odot$ denotes element-wise multiplication.

\paragraph{Masking reference patches}
To vary the complexity of the position prediction task, we mask a ratio $\eta$ of the reference patch representations $Z_{ref}$ that are visible to the query patch representations as in LOCA \cite{Caron2024loca}.

\paragraph{Patch cluster prediction}
To learn representations effective for pixel-level classification
(a fundamental part of semantic segmentation) without labels, we
generate pseudo-labels through clustering, following LOCA.
The pseudo-labels (soft cluster assignments) are obtained based on the similarity between (learnable) cluster prototypes $Q \in \mathbb{R}^{K \times \tilde{d}}$ and projected patch representations of the reference view $\tilde{Z} \in \mathbb{R}^{N_{ref} \times \tilde{d}}$. A patch $i$ in the query will thus have a pseudo-label
\begin{align*}
    y^{j} = \text{Sinkhorn-Knopp}\left(\text{softmax}\left(\tilde{Z}_{ref}^{j} \cdot Q / \tau \right)\right)    
\end{align*}
where $j = h(i)$ and $\tau$ is the temperature parameter controlling the sharpness of the softmax distribution. We use $\tau = 0.05$. $\tilde{Z}$ is a projection of $Z$ by a two-layer MLP. The Sinkhorn-Knopp algorithm is used to prevent the model from collapsing to a trivial solution \cite{Caron2024loca}. We minimise the objective
\begin{align}
    \frac{1}{|\Omega|} \sum_{j \in \Omega} \ell((Q^\top \tilde{Z}_q)_j, y_j) \label{cluster-loss}
\end{align}

As in LOCA \cite{Caron2024loca}, we regularise this loss with mean entropy maximisation to encourage the network to use all cluster prototypes.

The combined objective includes equations \ref{pos-pred-loss} and \ref{cluster-loss} with equal weighting. 

\paragraph{Training and Evaluation}
We pretrain our models and the baseline methods on the MMEarth
multimodal satellite imagery dataset \cite{Nedungadi2025mmearth}.
We use a portion of 300,000 samples from MMEarth to reduce pretraining time,
and use only the Sentinel 2, Sentinel 1 and Aster DEM modalities.
We pretrain our models using AdamW optimisation with learning rate
$6.25 \times 10^{-5}$, cosine scheduling, batch size 64, and weight
decay 0.1. Both our models and the baseline methods are pretrained
for 100 epochs. We train the baseline methods using their respective
public implementation source code. Evaluation is done by end-to-end
fine-tuning on the Sen1Floods11 flood mapping semantic segmentation
dataset \cite{Bonafilia2020Sen1Floods11}. We use a light decoder
with four transposed convolution layers and a final convolution
layer that outputs the segmentation logits to prevent the pretrained
weights from being dissipated by a heavy decoder. The reported
evaluation results are averaged over three runs.

\section{Experiments}
\paragraph{Channel grouping on Sentinel 2}
We compare the performance of pretraining on Sentinel 2 images
with and without channel grouping. Following SatMAE
\cite{Cong2022satmae}, we group the Sentinel 2 bands by similarity
of spatial resolution and wavelength as follows. (See Appendix
\ref{app:bands} for band details.)
\begin{itemize}
    \item RGB and NIR bands: $B2, B3, B4, B8$
    \item Red Edge bands 1 to 4: $B5, B6, B7, B8A$
    \item SWIR bands 1 and 2: $B11, B12$.
\end{itemize}
We denote this group \textit{``S2 Similarity"}. Results in Tab. \ref{tab:changrouping} show that channel grouping is important
when dealing with multispectral imagery, yielding a performance
increase when applied to the finetuning and pretraining stages.
In the pretraining stage, with channel grouping, a query patch
in a certain channel group, say, SWIR bands, predicts its position
in reference view comprising all the groups. We hypothesise that
this cross-group interaction challenges the model to extract and
relate the distinct information from each group, thus obtaining
richer aggregated information.

\begin{table}
  \caption{\textbf{Channel grouping on Sentinel 2.} IoU of
  flood class and mIoU on Sen1Floods11 with and without channel
  grouping Sentinel 2 images.}
  \label{tab:changrouping}
  \begin{tabular}{cccc}
    \toprule
    \multicolumn{2}{c}{Channel grouping.} & IoU (flood) & mIoU\\
    Pretraining & Finetuning & &\\
    \midrule
     & & 69.12 & 82.51\\
     & \checkmark & 73.06 & 84.75\\
     \checkmark & \checkmark & 73.90 & 85.24 \\
  \bottomrule
\end{tabular}
\end{table}

\paragraph{Group sampling} To manage the computational cost of
pretraining we randomly sample one group for each patch position
thus maintaining a constant sequence lengths.
Tab. \ref{tab:group_sampling} shows that, for the \textit{S2 Similar}
group setting, we get a $\times\, 4.2$ reduction in gigaflops at the
cost of a $-\, 0.48$ mIoU decrease. The \textit{Best} group setting
is the group setting that eventually yields the best performance (See paragraph \textit{`Adding DEM modality as a channel group'}).
It contains 6 groups, thus group sampling results in $\times\, 12.2$
reduction in gigaflops. Interestingly, the performance decrease is
only $-0.03$ mIoU.

All following experiments are performed with group sampling.

\begin{table}
    \caption{\textbf{Group sampling}. Effect of group sampling on
    computational cost and flood segmentation performance on
    Sen1Food11.}
    \centering
    \begin{tabular}{cclcc}
        \toprule
         Group setting & Group sampling  & Speedup & IoU (flood) & mIoU \\
        \midrule
        \multirow{2}{*}{\textit{S2 Similar}} &  & --- & 73.90 & 85.24 \\
        & \checkmark  & $\times\ 4.2$ & 73.08 & 84.76 \\
        \specialrule{0.2pt}{2pt}{2pt}
        \multirow{2}{*}{\textit{Best}} &  & --- & 72.86 & 84.64 \\
        & \checkmark  & $\times\ 12.2$ & 72.82 & 84.61 \\
        \bottomrule
    \end{tabular}
    \label{tab:group_sampling}
\end{table}

\paragraph{Adding Sentinel 1 as channel groups}
We add the Sentinel 1 modality using the channel group architecture.
We define new channel group settings that include Sentinel 1 bands as
follows.
\begin{itemize}
    \item \textit{S2+S1 Separate}: \textit{S2 Similar} + \{(A-VV, A-VH, D-VV, D-VH),
    (A-HH, A-HV, D-HH, D-HV)\}

    \item \textit{RGBN+S1 Separate}: \{(B2), (B3), (B4), (B8),
    (A-VV, A-VH, D-VV, D-VH), (A-HH, A-HV, D-HH, D-HV)\}

    \item \textit{S2+S1 Mixed}: \textit{S2 Similar} + \{(B1, A-VV, A-VH, D-VV, D-VH),
    (B1, A-HH, A-HV, D-HH, D-HV)\}
\end{itemize}
The results in Tab. \ref{tab:multimodal} show that adding Sentinel 1
bands increases performance as long as the two modalities are grouped
separately, as in the \textit{S2+S1 Similar} and \textit{RGBN+S1}
group settings. These settings encourage cross-modal interaction
since a query patch representation from one modality must predict its
position by attending to all the modalities. We hypothesise that this
cross-modal interaction teaches the model to extract and combine
information more effectively from the multimodal data. We also see
that adding the Sentinel 1 modality as separate channel groups makes
the pretraining task more challenging, resulting in $-25\%$ accuracy
reduction in the pretraining objective. Mixing the bands of the
different modalities, as in the \textit{S2+S1 Mixed} group setting,
does not improve performance. Mixing bands reduces the need for
cross-modal interaction as query patch representations have
information from all modalities and can rely on the convenient
modality to solve the pretext task. We also see that the pretext
task is not much harder in the mixed setting than in the single-modal
setting ($54.92\%$ vs. $60.50\%$ resp.)

\begin{table}
    \caption{Ablations study on adding the SAR modality using the
    channel group architecture}
    \centering
    \begin{tabular}{ccc|c}
        \toprule
         Group setting & IoU (flood) & mIoU & Pretraining objective (acc@1)  \\
        \midrule
         \textcolor{gray}{\textit{S2 Similar}} & \textcolor{gray}{73.08} & \textcolor{gray}{84.76} & \textcolor{gray}{60.5} \\
         \textit{S2+S1 Separate} & 73.68 & 84.87 & 35.33 \\
         \textit{RGBN+S1 Separate} & 73.38 & 85.10 & 30.93 \\
         \textit{S2+S1 Mixed} & 72.40 & 84.35 & 54.92 \\
         \bottomrule
    \end{tabular}
    \label{tab:multimodal}
\end{table}

\paragraph{Adding DEM modality as a channel group}\label{para:add_dem}
To add DEM using channel groups, we introduce two new channel group
settings:
\begin{itemize}
    \item \textit{S2 + S1 + DEM Separate}: \textit{S2 + S1 Separate} + \{(DEM)\}
    \item \textit{Best}: \{(B1, B2), (B3, B7), (B4, B8A), (B11), (DEM, A-VV, A-VH, D-VH), (A-HH, A-HV, D-VV, D-HH)\}
\end{itemize}
The \textit{``Best"} group setting is the one that yields the best
performance on Sen1Floods11. It separates the MSI and SAR modalities
but mixes DEM into SAR.

Tab. \ref{tab:adding_dem} shows that adding DEM as a separate channel
group makes the pretraining task more challenging, yielding a lower
position prediction accuracy (-16.53\%). Mixing DEM into the present
modalities makes the pretraining task relatively simple, resulting in
a small decrease in the position prediction accuracy (-0.13\%). This
shows that the strategy for incoporating modalities is a
hyperparameter that can be tuned to control the difficulty of the
pretraining task and improve transfer performance.

Interestingly, a reference masking ratio of $\eta = 100\%$ yields the
the best performance, showing that there is no need for the query
patches to `look' at the reference view representations. Therefore,
our scheme's complexity and computation cost can be reduced by not
including the cross-attention block.

\begin{table}
    \caption{\textbf{Adding DEM.} Effect of different strategies of
        adding a third modality on performance on Sen1Flood11}
    \centering
    \begin{tabular}{ccccc}
        \toprule
        Group setting & $\eta$ & IoU (flood) & mIoU & Pretraining objective (acc@1) \\
        \midrule
        \textcolor{gray}{\textit{S2+S1 Separate}} & \textcolor{gray}{80\%} & \textcolor{gray}{73.68} & \textcolor{gray}{84.87} & \textcolor{gray}{35.33} \\
        \specialrule{0.2pt}{2pt}{2pt}
        \multirow{2}{*}{\textit{S2 + S1 + DEM Separate}} & 80\% & 72.11 & 84.38 & 13.8\% \\
        & 100\% & 73.88 & 85.21 & 1.54\% \\
        \specialrule{0.2pt}{2pt}{2pt}
        \multirow{2}{*}{\textit{Best}} & 80\% & 72.44 & 84.35 & 35.20\% \\
        & 100\% & 74.52 & 85.52 & 1.57\% \\
        \bottomrule
    \end{tabular}
    \label{tab:adding_dem}
\end{table}

\paragraph{Same-group attention masking}
We experiment with same-group attention masking as a technique for
improving multimodal learning by encouraging cross-modal interaction.
Tab. \ref{tab:attn_masking} shows that same-group attention masking
significantly improves transfer performance ($+1.89$ IoU) when the
reference masking ratio is low ($\eta = 60\%$). However, increasing
the reference masking ratio to $\eta = 100\%$ results in a slight
decrease in performance ($-0.06$ mIoU). Same-group attention
masking makes the pretraining task more challenging in a way that
helps the model learn better representations, however, combining it
with the aggressive reference masking reduces its effect since there
are few reference representations to attend to and possibly makes
the pretraining task too challenging for the model to learn good
representations.

\begin{table}
    \caption{\textbf{Same-group attention masking.} Effect of
        same-group attention masking and reference masking on transfer
        performance on Sen1Floods11}
    \centering
    \begin{tabular}{ccccc}
        \toprule
        $\eta$  & Same-group atten. masking & IoU (flood) & mIoU & Pretraining objective (acc@1) \\
        \midrule
        \multirow{2}{*}{60\%} & & 71.99 & 84.07 & 53.15 \\
        & \checkmark & 73.88 & 85.21 & 44.11 \\
        \specialrule{0.2pt}{2pt}{2pt}
        \multirow{2}{*}{100\%} & & 74.62 & 85.52 & 1.57 \\
        & \checkmark & 74.56 & 85.49 & 1.57 \\
        \bottomrule
    \end{tabular}
    \label{tab:attn_masking}
\end{table}

\paragraph{Patch cluster prediction}
Tab. \ref{tab:cluster_loss} shows that including the patch cluster
prediction task significantly positively affects transfer performance
($+1.77$ mIoU).

\begin{table}
    \caption{\textbf{Patch cluster prediction.} Effect of including
        the patch cluster prediction task.}
    \centering
    \begin{tabular}{ccc}
        \toprule
        Cluster loss & IoU (flood) & mIoU \\
        \midrule
        \checkmark & 73.88 & 85.21 \\
        & 72.11 & 84.06 \\
        \bottomrule
    \end{tabular}
    \label{tab:cluster_loss}
\end{table}

\paragraph{Comparison with other satellite imagery SSL pretraining schemes}
We compare our pretraining scheme to other popular schemes. We pretrain
ViT-Small encoders (or ConvNext-T encoder for the MMEarth scheme
\cite{Nedungadi2025mmearth}) for 100 epochs on the MMEarth dataset using
their publicly accessible implementation source code. For the MMEarth
scheme, we pretrain using Sentinel 1 and Sentinel 2 modalities only.
Evaluation is done through end-to-end fine-tuning using a light decoder
(4 transponse convolution layers plus a final pixel-level classification
convolution layer.) We report results from a single finetuning run of the
schemes.

Tab. \ref{tab:comparison} shows that our adopted LOCA method does
significantly better than the other methods on satellite imagery semantic
segmentation on the Sen1Floods11 dataset.

\begin{table}
    \caption{\textbf{Comparison with other SSL pretraining schemes on Sen1Floods11}}
    \centering
    \begin{tabular}{llcc}
        \toprule
        Scheme & Encoder & IoU (flood) & mIoU \\
        \midrule
        Satellite LOCA (ours) & ViT-Small & 74.62 & 85.49 \\
        MMEarth \cite{Nedungadi2025mmearth}  & ConvNext-T & 68.92 & 82.34 \\
        ScaleMAE \cite{Reed2023scalemae} & ViT-Small & 68.85 & 82.29 \\
        SatMAE++ \cite{Noman2024Satmaepp} & ViT-Small & 67.37 & 81.47 \\
        SatMAE \cite{Cong2022satmae} & ViT-Small & 65.28 & 80.56 \\
        \bottomrule
    \end{tabular}
    \label{tab:comparison}
\end{table}

\section{Conclusion}
We adapt LOCA, a position prediction self-supervised learning method, for
multimodal satellite imagery semantic segmentation. Our key contributions
include extending channel grouping to handle multimodal data, introducing
same-group attention masking to encourage cross-modal interaction, and
using group sampling to maintain computational efficiency during pretraining.
Experimental results on Sen1Floods11 show that our approach significantly
outperforms existing reconstruction-based self-supervised methods for
satellite imagery. Future work could explore incorporating scale-invariance
mechanisms in the pretraining as in ScaleMAE \cite{Reed2023scalemae},
exploiting the temporal dimension of satellite data, extending to additional
modalities, and evaluating transfer learning on more diverse downstream tasks.

% \begin{acks}
% \end{acks}

\bibliographystyle{ACM-Reference-Format}
\bibliography{ref}

\appendix
\section{Modalities and Bands} \label{app:bands}
Tab. \ref{tab:bands_ref} lists the bands of the modalities used in
this work.

\begin{table}[H]
    \caption{The modalities and bands used, with the codes used
        to reference them.}
    \centering
    \begin{tabular}{p{2cm}llc}
        \toprule
         Modality & Code & Name & Spatial Resolution (metres)\\
        \midrule
        \multirow{13}{2cm}{Multspectral Satellite Imagery (Sentinel 2)} & B1 & Ultra-blue & 60\\
        & B2 & Blue & 10 \\
        & B3 & Green & 10 \\
        & B4 & Red & 10 \\
        & B5 & Red edge 1 & 20 \\
        & B6 & Red edge 2 & 20 \\
        & B7 & Red edge 3 & 20 \\
        & B8 & Near-infrared & 10 \\
        & B8A & Red edge 4 & 20 \\
        & B9 & Water vapour & 60 \\
        & B10 & Cirrus & 60 \\
        & B11 & Shortwave-infrared 1 & 20 \\
        & B12 & Shortwave-infrared 2 & 20 \\
        \specialrule{0.2pt}{2pt}{2pt}
        \multirow{8}{2cm}{Synthetic Aperture Radar (Sentinel 1)} & A-VV & Ascending orbit VV & 10 \\
        & A-VH & Ascending orbit VH & 10 \\
        & A-HH & Ascending orbit HH & 10 \\
        & A-HV & Ascending orbit HV & 10 \\
        & D-VV & Descending orbit VV & 10 \\
        & D-VH & Descending orbit VH & 10 \\
        & D-HH & Descending orbit HH & 10 \\
        & D-HV & Descending orbit HV & 10 \\
        \specialrule{0.2pt}{2pt}{2pt}
        Digital elevation model & DEM & Elevation & 30 \\
        \bottomrule
    \end{tabular}
    
    \label{tab:bands_ref}
\end{table}

\end{document}